# Improve Sentence Alignment by Divide-and-conquer

Wu Zhang
The University of Hong Kong
zhangwu@hku.hk

**Abstract**

In this paper, we introduce a divide-and-conquer algorithm to improve sentence alignment speed. We utilize external bilingual sentence embeddings to find accurate hard delimiters for the parallel texts to be aligned. We use Monte Carlo simulation to show experimentally that using this divide-and-conquer algorithm, we can turn any quadratic time complexity sentence alignment algorithm into an algorithm with average time complexity of $O(NlogN)$. On a standard OCR-generated dataset, our method improves the Bleualign baseline by 3 $F_1$ points. Besides, when computational resources are restricted, our algorithm is faster than Vecalign in practice.

## 1 Introduction

Sentence alignment is the task of sequentially finding all minimal paired sentences in two parallel texts, where minimal means that it cannot be sequentially split into two paired sentences and paired means that sentences on each side are translations of each other. As in most literature, we assume that there are no crossing alignments but local sentence reordering within an alignment is allowed.

Before large open-source parallel corpus is available, sentence alignment is usually the first step towards building a statistical machine translation (SMT) system. With the release of many standard machine translation corpus, sentence alignment gets less and less attention recently. Still, sentence alignment has its application in building para-crawl pipelines (Varga et al., 2007; Bañón et at., 2020) and building translation memory in computer assisted translation systems.

Sentence alignment is done by first defining a scoring function for alignments, and then using dynamic programming (DP; Bellman, 1953) to maximize or minimize a global alignment score. The time complexity of DP is quadratic.

In this paper, we introduce a divide-and-conquer algorithm aimed at improving sentence alignment speed and quality for existing algorithms. Our contributions are as follows: (1) we propose a method to find accurate hard delimiters for the parallel texts to be aligned using external bilingual sentence embeddings, (2) we show experimentally that by using divide-and-conquer, a quadratic time complexity algorithm can achieve an average time complexity of $O(NlogN)$, and (3) in a low computational resources scenario, our algorithm is usually faster than Vecalign (Thompson and Koehn, 2019) and can achieve comparable or even better accuracy than Vecalign for easy-to-align parallel texts.

Our algorithm improves the Bleualign (Sennrich and Volk, 2010) baseline by 3 $F_1$ points on a standard OCR-generated dataset which indicates that our method can improve accuracy too.

## 2 Related Work

Brown et al. (1991) and Gale and Church (1993) used scoring functions based solely on sentence length (the former counts tokens while the latter counts characters). These scoring functions performed quite well on easy-to-align corpus such as the English-French Hansard corpus. Chen (1993) introduced a method to optimize word translation probability in the corpus while Wu (1994) used an external bilingual lexicon to align the corpus. Later, a hybrid method was introduced to utilize both length and lexical information (Moore, 2002). Bleualign (Sennrich and Volk, 2010) used SMT models to translate sentences in one language into the other language and used a revised sentence level BLEU (Papineni et al., 2002) as the scoring function. Later, Bleualign is improved by using a coverage-based method (Gomes and Lopes, 2016). Previous efforts on improving sentence alignment speed have been made either through searching near diagonal or first finding high-confidence 1-to-1 alignments and then searching near the diagonal path passing through them. Vecalign (Thompson and Koehn, 2019) defined a scoring function using purely bilingual sentence embeddings and approximated the dynamic programming algorithm in linear time and space (Salvador and Chan, 2007). In practice, we found that Vecalign spent over 80% (over 90% for less than 3000 sentence pairs) of the time in computing sentence embeddings. An algorithm that embeds fewer sentences than Vecalign would be desirable to improve the overall speed.

## 3 Method

Inspired by Vecalign, we use external bilingual sentence embeddings to find high-confidence hard delimiters. Any sentence alignment algorithm can be applied in parallel to each chunk of the parallel texts segmented by hard delimiters.

### 3.1 Definition of hard delimiter

Any 1-to-1 alignments can be used as hard delimiters. But for practical reasons, we define hard delimiters in this paper as those 1-to-1 alignments which are surrounded by 1-to-1 alignments.

Formally, assume that sentences in the two parallel texts can be enumerated as $\{x_i\}_{i=1}^{N}$ and $\{y_i\}_{i=1}^{M}$, respectively. Assuming we know the gold-standard alignments, we define (true) hard delimiters as the set of all 1-to-1 alignments $(x_i, y_j)$ such that $(x_{i-1}, y_{j-1})$ and $(x_{i+1}, y_{j+1})$ are both 1-to-1 alignments.

We have two reasons for giving this definition. On one hand, this definition can reduce search errors by avoiding getting 1-to-1 alignments which should be part of a 1-to-many or many-to-many alignments. On the other hand, consider the case when we need to decide whether two identical numbers should be aligned (there is a possibility that there are the same numbers before and/or after them). We will be more confident that they should be aligned if their surrounding sentences are aligned.

### 3.2 Time complexity using hard delimiters

Assume that the parallel texts we are about to align have $N$ and $M$ sentences on each side. Assume that we have found $k$ hard delimiters and segmented the parallel texts into $k+1$ smaller parallel texts (we will call them **chunks**). Assume that the $i$-th chunk has $N_i$ and $M_i$ sentences on each side. Then for a quadratic time complexity algorithm, the time complexity to align all these chunks is

$$\sum_{i=0}^{k} O(N_i \times M_i) \leq \max_{0 \leq i \leq k} N_i \times \sum_{i=0}^{k} O(M_i)$$
$$= \max_{0 \leq i \leq k} N_i \times O(M)$$

### 3.3 Expected maximum chunk size

Assume that the gold-standard alignments contain $n$ total alignments and the ratio of 1-to-1 alignments is $r$. In the worst case, the maximum chunk size (evaluated in the number of alignments) is $n \times (1-r) + 1$ and a quadratic time complexity algorithm will still have quadratic time complexity even if we can find all hard delimiters. However, this case is very rare, and in most cases, the maximum chunk size should be very small compared to the magnitude of $n$. The expected maximum chunk size is

$$\sum_{a \in A} \frac{1}{\binom{n}{nr}} \times G(a),$$

where $A$ is the set of all possible alignment combinations and $G(a)$ is the maximum chunk size of an alignment arrangement $a$.

The expected maximum chunk size is hard to compute. We run Monte Carlo experiments (Rubinstein, 1981) to estimate the expected maximum chunk size for a given $n$ and $r$.

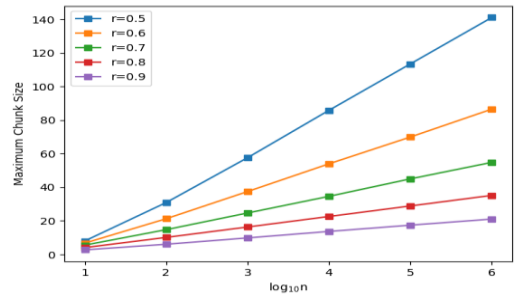

Figure 1: Expected maximum chunk size

From Figure 1, we can see that for a given $r$, the expected maximal chunk size is roughly linear in $\log n$ which indicates that the expected maximum chunk size is $O(\log n)$.

### 3.4 Finding hard delimiters using bilingual sentence embeddings

Bilingual sentence embeddings have been successful in bitext mining (Guo et al., 2018) and filtering out non-parallel sentences (Hassan et al., 2018; Chaudhary et al., 2019). Methods like margin-based bitext mining (Artetxe and Schwenk, 2019) have quadratic time complexity. Vecalign used bilingual sentence embeddings to define a new scoring function and applied a DP approximation method (Salvador and Chan, 2007) that runs in linear time and space. Vecalign computes the embeddings for $m \times (N + M)$ sentences, where $m$ is the hyperparameter "alignment_max_size" of Vecalign which defines the maximum number of total sentences that an alignment can contain and $N$ and $M$ are the number of sentences on each side of the parallel texts. We found that in a low computational resource scenario, sentence embedding time is the bottleneck of Vecalign for an alignment task. Hence, we can reduce the number of sentences to be embedded to reduce the total alignment time. We only want to find 1-to-1 alignments, so we can set $m = 2$. However, we found that this will reduce the accuracy of finding hard delimiters.

Theoretically, we can use k-d tree to do global nearest neighborhood search. Building a balanced k-d tree takes $O(kN\log N)$ time (Brown, 2015), where $k$ is the sentence embedding dimension and $N$ is the number of sentences (assume $N = M$). Querying for the nearest neighbor can be done in $O(\log N)$ time. Hence, we can find all pairs of most similar sentences (note that when all vectors are normalized, then the smallest L2 neighborhood is equivalent to the biggest cosine similarity) in $O(N\log N)$ time when $k$ is fixed. We can regard those pairs of sentences as our candidates for 1-to-1 alignments (they may not be in order). Then we find the longest in order alignments from them. The time complexity of this step is $O(N\log N)$. Thus, the total time complexity of this strategy is $O(N\log N)$. This finishes the proof of the average time complexity of our algorithm.

Note that with similarity search tools like faiss[1], doing bitext mining using GPU for 100,000 sentences only takes several seconds. In practice, we will use margin-based bitext mining methods (Artetxe and Schwenk, 2019) to search for our 1-to-1 alignment candidates.

### 3.5 DP approximation within a chunk

In some bad cases, the maximum chunk size is very big. To deal with these cases, we can either try to find hard delimiters within each chunk (this time, search scope is smaller than the global search) or try to find high-confidence monotonic 1-to-1 alignments and search near the diagonal path passing through these alignments during DP. A better strategy is to combine both methods. We can first iteratively find all possible hard delimiters, and then for big chunks, we find high-confidence monotonic 1-to-1 alignments using bitext mining technique and then search near the diagonal path passing through these alignments during DP.

## 4 Experiments and Results

### 4.1 Metrics and hardware settings

We use the *strict* precision, recall, and $F_1$ defined in Bleualign to evaluate alignment accuracy. For hard delimiters, we compute the precision, recall and $F_1$ score using the hard delimiters found by our algorithm against the true hard delimiters computed from the gold-standard alignments. We used a machine with 64GB RAM, 16 Intel(R) Core (TM) i7-10700K CPU, and an Nvidia GeForce RTX 2080 Ti GPU.

### 4.2 OCR-generated dataset accuracy

We use LASER[2] as our sentence embedding tool. We test the accuracy of finding hard delimiters on the OCR-generated dataset released by Bleualign. This dataset consists of manually aligned German-French articles getting from the Text+Berg corpus (Volk et al., 2010). It contains a development set of 1 article pair and a test set of 7 article pairs. Like Vecalign, hyperparameters were chosen to optimize the $F_1$ of finding hard delimiters on the development set. The only hyperparameter of our algorithm is a cosine similarity filtering threshold for filtering 1-to-1 alignments found using bitext mining. It is set to 0.6 across all experiments afterward.

On the 7 test article pairs, precision, recall, and $F_1$ of finding hard delimiters range from 0.93-1, 0.45-0.81, and 0.61-0.87, respectively. We computed them separately because our algorithm works on article level. We proceeded to align the seven article pairs using length-based algorithm

---

[1] https://github.com/facebookresearch/faiss

[2] https://github.com/facebookresearch/LASER

(Gale and Church, 1993), Bleualign (Sennrich and Volk, 2010), and Vecalign (Thompson and Koehn, 2019). We choose them because on this dataset, their performance is relatively high, their results are easy to replicate, and they work on single parallel texts level.

| Algorithm | P | R | $F_1$ |
|---|---|---|---|
| Gale & Church | 0.71 | 0.72 | 0.72 |
| DAC + Gale & Church | 0.80 | 0.80 | 0.80 |
| Bleualign | 0.83 | 0.78 | 0.81 |
| DAC + Bleualign | 0.83 | 0.85 | 0.84 |
| Vecalign | **0.89** | **0.90** | **0.90** |
| DAC + Vecalign | 0.88 | 0.89 | 0.88 |

Table 1: Strict precision (P), recall (R), and $F_1$ on Bleualign test set. "DAC" is the abbreviation for "divide-and-conquer".

Results are shown in Table 1. Our algorithm improves the Bleualign baseline by 3 $F_1$ points. The performance of DAC+Vecalign is worse than Vecalign which indicates that Vecalign itself can find very accurate 1-to-1 alignments.

### 4.3 Speed test

Length-based algorithms (Gale and Church, 1993; Brown et al., 1991) are very fast. Its speed can beat Vecalign when the size of the parallel texts is not too big. Here we compare the time cost of Vecalign with a lexical-based algorithm strengthened with our divide-and-conquer algorithm.

We use the scoring function defined in Gargantua (Braune and Fraser, 2010) with the length-based term being removed. Note that Gurgantua's scoring function is a revised version of Moore's (Moore, 2002) scoring function. We trained a fastalign[3] (Dyer et al., 2013) model on more than 20 million EN-ZH sentence pairs to get a lexical translation table. During DP, we search for all 1-to-many and many-to-1 beads up to 1-to-5 and 5-to-1 with an additional 2-to-2 bead.

We have three test article pairs. The first one is highly symmetric and less noisy while the latter two are pdf converted articles and are highly asymmetric and noisier. They contain 882/871, 3069/3407, and 26037/24725 English/Chinese sentences, respectively. We manually aligned the first pair to test the accuracy of our algorithm. The latter two serve the purpose of testing the speed of our algorithm in some bad cases.

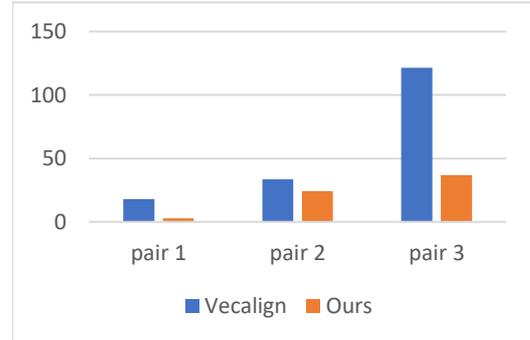

Figure 2: Total time cost (in seconds) for 3 test sets.

The time cost of Vecalign and our algorithm are shown in Figure 2. Sentence embedding time is added to the total time. We fixed the "alignment_max_size" of Vecalign to be 6 during testing. On the three pairs of articles, Vecalign took 17.9/33.6/121.4 seconds, respectively. Our divide-and-conquer plus lexical-based algorithm took 2.8/24.2/36.9 seconds, respectively. Sentence embedding time for Vecalign on the last test pair is about 103s which is roughly in line with the time of embedding 32k sentences (~120s) reported in Vecalign. Our algorithm slightly outperformed Vecalign on the first test pair with a strict precision/recall/F1 of 0.992/0.989/0.990. Vecalign got 0.979/0.990/0.984 on this pair. Bitext mining method got 0.745/0.865/0.800.

## 5 Conclusion

We introduced a divide-and-conquer algorithm that utilizes sentence embedding similarity to find high-quality hard delimiters. Our algorithm can improve the speed (and quality) of quadratic time complexity sentence alignment algorithms and make them faster than the linear time algorithm Vecalign in practice.

## Acknowledgments

We thank the authors of Bleualign and Vecalign for sharing their code and for making their results easy to replicate.

---
[3] https://github.com/clab/fast_align